\documentclass{article}
\PassOptionsToPackage{numbers,sort&compress}{natbib}
\usepackage[preprint]{neurips_2026}
\usepackage[utf8]{inputenc}
\usepackage[T1]{fontenc}
\usepackage[protrusion=true,expansion=false]{microtype}

\usepackage{amsmath,amssymb,amsfonts,booktabs,tabularx,array,graphicx,xcolor}
\usepackage{tikz}
\usetikzlibrary{positioning,arrows.meta,calc}
\usepackage{pgfplots}
\pgfplotsset{compat=1.18}
\usepgfplotslibrary{groupplots}
\usepackage{caption}
\usepackage{enumitem}
\usepackage{needspace}
\setlist{itemsep=2pt,topsep=4pt}
\usepackage[section]{placeins}
\usepackage{xurl}
\usepackage{hyperref}
\hypersetup{colorlinks=true,linkcolor=blue!45!black,citecolor=blue!45!black,urlcolor=blue!45!black,
 pdftitle={EHR2Trace: Auditable EHR Data Infrastructure for Patient World Models and Clinical Agents},
 pdfsubject={Auditable EHR data infrastructure for patient world models, clinical agents, and offline reinforcement learning}}
\newcolumntype{Y}{>{\raggedright\arraybackslash}X}
\makeatletter\newif\ifanon\if@anonymous\anontrue\fi\makeatother
\newcommand{\fitwidth}[1]{\resizebox{\linewidth}{!}{#1}}
\newcommand{\mimicShards}{364{,}673}
\newcommand{\leakSubjects}{131{,}007}
\newcommand{\leakDeaths}{4{,}185}
\newcommand{\leakPrevalence}{3.2\%}
\newcommand{\leakCleanAuroc}{0.829}
\newcommand{\leakDatedAuroc}{0.965}

\newcommand{\leakCleanAuprc}{0.222}
\newcommand{\leakDatedAuprc}{0.654}

\newcommand{\DqdBlindWord}{Eight}

\newcommand{\dqdReachingWord}{nine}

\newcommand{\reproConfigs}{4}

\newcommand{\reproArtifacts}{60}

\newcommand{\faultCount}{28}
\newcommand{\faultsDetected}{28}

\newcommand{\checksAddedAfterMiss}{21}

\newcommand{\mimicCanonicalMinutes}{191}
\newcommand{\mimicCanonicalPeakGb}{167.2}
\newcommand{\mimicIngestMinutes}{111}
\newcommand{\mimicMedsMinutes}{128}
\newcommand{\mimicOmopMinutes}{47}
\newcommand{\cmpSubjects}{100}
\newcommand{\cmpOurEvents}{971{,}545}
\newcommand{\cmpLineageLinks}{972{,}567}
\newcommand{\cmpAvailabilityShare}{71\%}
\newcommand{\cmpBaselineDrugTypes}{one}
\newcommand{\cmpBaselineMedRecoverable}{99.8\%}
\newcommand{\cmpBaselineMedsEvents}{916{,}166}
\newcommand{\cmpBaselineOmopRows}{423{,}057}
\newcommand{\ReducedCheckCount}{34}
\newcommand{\ReducedFaultsDetected}{13}
\newcommand{\DqdFaultsDetected}{five}
\newcommand{\leakDatedGain}{0.136}
\newcommand{\leakDatedGainCi}{[0.116, 0.157]}
\newcommand{\leakResamples}{2{,}000}

\newcommand{\leakXferCleanAuroc}{0.829}
\newcommand{\leakXferDatedAuroc}{0.965}
\newcommand{\leakXferCrossAuroc}{0.642}
\newcommand{\leakXferInflation}{0.323}
\newcommand{\leakXferInflationCi}{[0.295, 0.352]}
\newcommand{\leakXferDamage}{0.188}
\newcommand{\leakXferDamageCi}{[0.161, 0.215]}
\newcommand{\leakXferDamageMin}{0.169}
\newcommand{\leakXferDamageMax}{0.188}

\newcommand{\leakXferHorizons}{4}
\newcommand{\leakXferCrossAuprc}{0.071}
\newcommand{\leakXferBDamageMax}{0.012}
\newcommand{\leakXferReverseAuroc}{0.721}

\newcommand{\currentCheckCount}{55}

\newcommand{\totalCanonicalMillions}{846.4}
\newcommand{\drugGold}{139}
\newcommand{\drugResolved}{129}

\newcommand{\drugAgreement}{93.8}
\newcommand{\placeOfOccurrenceRows}{75{,}325}
\newcommand{\atrialRateRows}{450{,}357}
\newcommand{\loincCommunity}{1{,}397}

\newcommand{\mimicSources}{33}

\newcommand{\ctpeVitalTimeless}{0}

\newcommand{\mimicVitalTimeless}{0}

\newcommand{\mimicNonMedicationOrders}{28{,}916{,}331}
\newcommand{\mimicPyxisEvents}{1{,}189{,}422}
\newcommand{\mimicEmarDose}{33{,}817{,}134}
\newcommand{\mimicEmarRoute}{13{,}436{,}489}
\newcommand{\mimicDispenses}{18{,}968{,}537}
\newcommand{\ctpeAssumedAvailabilityPct}{51.0}
\newcommand{\mimicAssumedAvailabilityPct}{11.3}
\newcommand{\cuVitalTimeless}{0}

\newcommand{\cuAssumedAvailabilityPct}{99.8}

\definecolor{ehrteal}{RGB}{13,110,102}
\definecolor{ehrblue}{RGB}{31,86,140}
\definecolor{ehrorange}{RGB}{191,87,32}
\definecolor{ehrslate}{RGB}{70,78,86}

\newlength{\figunit}

\tikzset{
  figbase/.style={
    font=\footnotesize,
    inner sep=0pt,
    outer sep=0pt,
    line width=0.5pt,
  },
  panel/.style={execute at begin node=\nohyph, 
    draw=ehrblue!55, fill=ehrblue!5, rounded corners=0.6mm,
    line width=0.6pt, align=left, inner sep=1.6\figunit,
  },
  target/.style={execute at begin node=\nohyph, 
    draw=ehrteal!60, fill=ehrteal!6, rounded corners=0.6mm,
    line width=0.6pt, align=left, inner sep=1.4\figunit,
  },
  band/.style={execute at begin node=\nohyph, 
    draw=none, fill=black!5, rounded corners=0.6mm,
    align=left, inner sep=1.4\figunit,
  },
  proposed/.style={execute at begin node=\nohyph, 
    draw=ehrorange!75, fill=ehrorange!4, rounded corners=0.6mm,
    dash pattern=on 1.1mm off 0.9mm, line width=0.6pt,
    align=left, inner sep=1.6\figunit,
  },
  chip/.style={execute at begin node=\nohyph, 
    draw=ehrteal!35, fill=white, rounded corners=0.4mm,
    line width=0.4pt, align=center, inner sep=0.9\figunit,
  },
  ptitle/.style={font=\small\bfseries, color=ehrblue},
  ttitle/.style={font=\small\bfseries, color=ehrteal},
  otitle/.style={font=\small\bfseries, color=ehrorange},
  btitle/.style={font=\footnotesize\bfseries, color=ehrslate},
  bodytext/.style={font=\footnotesize, color=black!78, align=left,
                   execute at begin node=\nohyph},
  minitext/.style={font=\scriptsize, color=black!70, align=left,
                   execute at begin node=\nohyph},
  flow/.style={draw=ehrslate, line width=0.7pt, -{Stealth[length=1.5mm,width=1.2mm]}},
  feed/.style={draw=black!45, line width=0.5pt, dash pattern=on 0.8mm off 0.7mm,
               -{Stealth[length=1.3mm,width=1.1mm]}},
  demand/.style={draw=ehrorange!80, line width=0.6pt, dash pattern=on 1.1mm off 0.9mm,
                 -{Stealth[length=1.4mm,width=1.2mm]}},
}

\newcommand{\figsep}{\,\textperiodcentered\,}

\tikzset{
  cellbase/.style={minimum size=3.1\figunit, inner sep=0pt, outer sep=0pt,
                   rounded corners=0.25mm},
  cellhit/.style={cellbase, fill=ehrteal, draw=none},
  cellmiss/.style={cellbase, fill=ehrorange, draw=none},
  cellscope/.style={cellbase, fill=black!11, draw=none},
  cellnr/.style={cellbase, fill=white, draw=black!40, dashed, line width=0.3pt},
}

\pgfplotsset{
  leakpanel/.style={
    font=\footnotesize,
    title style={font=\footnotesize\bfseries, color=ehrslate, yshift=0.6mm},
    axis line style={black!40, line width=0.5pt},
    tick style={black!40, line width=0.5pt},
    tick align=outside, xmin=0.65, xmax=4.35,
    xlabel near ticks, xlabel style={font=\footnotesize, yshift=-1.7mm},
    enlarge x limits=false,
    grid=major, grid style={black!13, line width=0.35pt},
    tick label style={color=black!65},
    label style={color=black!70},
    scaled y ticks=false, yticklabel style={/pgf/number format/fixed},
    legend style={draw=none, fill=none, font=\footnotesize, text=black!70,
                  column sep=6mm, /tikz/every even column/.append style={column sep=4mm}},
  },
}
\tikzset{
  armfilter/.style={color=ehrblue, line width=1.0pt, mark=*, mark size=1.4pt,
                    mark options={fill=ehrblue, draw=ehrblue}},
  armignored/.style={color=ehrteal, line width=1.0pt, dash pattern=on 1.6mm off 1.1mm,
                     mark=diamond*, mark size=1.8pt,
                     mark options={solid, fill=white, draw=ehrteal, line width=0.7pt}},
  armdated/.style={color=ehrorange, line width=1.0pt, mark=square*, mark size=1.4pt,
                   mark options={fill=ehrorange, draw=ehrorange}},
}

\newcommand{\figband}[1]{{\footnotesize\bfseries\color{ehrslate}#1}}
\newcommand{\figbody}[1]{{\footnotesize\color{black!78}\nohyph #1}}
\newcommand{\figmini}[1]{{\scriptsize\color{black!70}\nohyph #1}}

\newcommand{\nohyph}{\hyphenpenalty=10000\exhyphenpenalty=10000\relax}

\tikzset{
  stage/.style={execute at begin node=\nohyph, draw=black!32, fill=white, rounded corners=0.6mm,
                line width=0.6pt, align=left, inner sep=1.3\figunit},
  stagework/.style={stage, draw=ehrteal, fill=ehrteal!8, line width=1.2pt},
  stagefuture/.style={stage, draw=ehrorange!75,
                      dash pattern=on 1.0mm off 0.8mm},
  zoom/.style={draw=ehrteal!50, line width=0.5pt},
  detailbg/.style={fill=ehrteal!4, rounded corners=1mm},
}
\newcommand{\figstage}[2]{{\footnotesize\bfseries\color{#1}#2}}

\tikzset{
  iconbase/.style={line width=0.65pt, line cap=round, line join=round},
  ic sources/.pic={
    \draw[rounded corners=0.3mm] (-2.7,-1.9) rectangle (0.7,2.7);
    \draw[rounded corners=0.3mm, fill=white] (-0.9,-2.7) rectangle (2.5,1.9);
    \draw (-0.1,1.0) -- (1.7,1.0);
    \draw (-0.1,-0.2) -- (1.7,-0.2);
    \draw (-0.1,-1.4) -- (1.0,-1.4);
  },
  ic layers/.pic={
    \draw[rounded corners=0.3mm] (-2.8,1.5) rectangle (1.6,2.8);
    \draw[rounded corners=0.3mm] (-2.8,-0.4) rectangle (1.6,0.9);
    \draw[rounded corners=0.3mm] (-2.8,-2.3) rectangle (1.6,-1.0);
    \draw[line width=0.9pt] (0.9,-1.5) -- (1.9,-2.5) -- (3.3,-0.4);
  },
  ic path/.pic={
    \draw (-2.8,-1.9) .. controls (-1.2,-2.4) and (-0.6,0.9) .. (0.6,0.6)
                      .. controls (1.6,0.35) and (1.7,2.0) .. (2.7,2.3);
    \fill (-2.8,-1.9) circle (0.44);
    \fill (0.6,0.6) circle (0.44);
    \fill (2.7,2.3) circle (0.44);
  },
  ic trajectory/.pic={
    \draw (-2.9,-2.5) -- (2.9,-2.5);
    \draw (-1.9,-2.5) -- (-1.9,-0.9);  \fill (-1.9,-0.9) circle (0.44);
    \draw (0.1,-2.5) -- (0.1,0.7);     \fill (0.1,0.7) circle (0.44);
    \draw (2.1,-2.5) -- (2.1,2.3);     \fill (2.1,2.3) circle (0.44);
  },
  ic agent/.pic={
    \draw (0,2.1) -- (-2.3,-1.5) -- (2.3,-1.5) -- cycle;
    \draw (0,2.1) -- (0,-1.5);
    \fill (0,2.1) circle (0.5);
    \fill (-2.3,-1.5) circle (0.5);
    \fill (2.3,-1.5) circle (0.5);
  },
  ic evaluation/.pic={
    \draw (-2.9,-2.5) -- (2.9,-2.5);
    \fill (-2.2,-2.5) rectangle (-1.1,-0.5);
    \fill (-0.5,-2.5) rectangle (0.6,1.1);
    \fill (1.2,-2.5) rectangle (2.3,-0.1);
  },
  ic table/.pic={
    \draw[rounded corners=0.3mm] (-2.7,-2.1) rectangle (2.7,2.1);
    \draw (-2.7,0.9) -- (2.7,0.9);
    \draw (-2.7,-0.5) -- (2.7,-0.5);
    \draw (-0.9,2.1) -- (-0.9,-2.1);
    \draw (0.9,2.1) -- (0.9,-2.1);
  },
  ic tokens/.pic={
    \draw[rounded corners=0.25mm] (-2.9,-0.5) rectangle (-1.5,1.7);
    \draw[rounded corners=0.25mm, fill=none] (-1.1,-0.5) rectangle (0.3,1.7);
    \draw[rounded corners=0.25mm] (0.7,-0.5) rectangle (2.1,1.7);
    \draw[rounded corners=0.25mm, fill=none] (2.5,-0.5) rectangle (3.1,1.7);
    \draw (-2.9,-1.7) -- (3.1,-1.7);
  },
}

\title{EHR2Trace: Auditable EHR Data Infrastructure for Patient World Models and Clinical Agents}

\author{%
  Xinye Yang \\
  Department of Radiology \\
  University of Colorado Anschutz School of Medicine \\
  Aurora, CO, USA \\
  \texttt{yangxinye2001@gmail.com} \\
  \And
  Yuli Wang \\
  Department of Radiology \\
  Johns Hopkins University School of Medicine \\
  Baltimore, MD, USA \\
  \texttt{ywang687@jhmi.edu} \\
  \And
  Cheng Ting Lin \\
  Department of Radiology \\
  Johns Hopkins University School of Medicine \\
  Baltimore, MD, USA \\
  \texttt{clin97@jhmi.edu} \\
  \And
  Harrison Bai\thanks{Corresponding author.} \\
  Department of Radiology \\
  University of Colorado Anschutz School of Medicine \\
  Aurora, CO, USA \\
  \texttt{harrison.bai@cuanschutz.edu} \\
}

\begin{document}

\maketitle

\begin{abstract}
Patient world models and clinical agents require histories that distinguish clinical events, treatment actions, and the information available at each decision. We present EHR2Trace, a configurable system that converts heterogeneous electronic health records (EHRs) into source-linked patient events and exports them to OMOP and MEDS. The system separates event time from information availability, distinguishes medication orders, dispensing, and administration, and validates saved outputs against source and audit records. Across three clinical datasets, EHR2Trace converted \totalCanonicalMillions{} million events and detected all \faultsDetected{} faults in an injected catalogue. Repeated builds produced \reproArtifacts{} byte-identical artifacts across \reproConfigs{} execution configurations on synthetic data. In a retrospective mortality-proxy experiment with availability-filtered test histories, the model trained with availability filtering achieved six-hour held-out AUROC \leakXferCleanAuroc{}, compared with \leakXferCrossAuroc{} for the model trained with backdated diagnoses. Cross-rule evaluation also exposed inflated AUROC of \leakDatedAuroc{} when backdated histories were used for both training and testing. EHR2Trace delivers reusable, auditable data infrastructure for constructing patient histories and quantifying the effects of conversion choices on downstream models.
\end{abstract}

\section{Introduction}

Patient world models and clinical agents require longitudinal records that distinguish a patient's condition, the care provided, and the information available at each decision. World models use these records to predict changes in patient state following clinical actions~\citep{ethos2024,ehrworld2026}, while clinical agents retrieve information and perform tasks in EHR environments~\citep{physicianbench2026}. Preserving these distinctions during conversion establishes a reliable basis for model training and evaluation.

EHR conversion involves decisions about the meaning and timing of clinical records. Medication orders, dispensing, and administration represent different stages of care. Similarly, the time of an event may differ from the time at which it becomes available to a clinician or a model. Conflating these distinctions introduces future information into predictions or represents a dispensed drug as an administered treatment, even when the exported data satisfy schema requirements. Temporal leakage inflates measured performance~\citep{kapoor2023}. For offline reinforcement learning, treatment histories also require explicit definitions of actions, rewards, and episodes, together with methods for addressing confounding in observational data~\citep{gottesman2019}.

\begin{figure}[t]
\centering
%
\setlength{\figunit}{0.01\linewidth}%
\begin{tikzpicture}[figbase, x=\figunit, y=\figunit]

\node[minitext, anchor=north west] at (0,1.6)
  {\figstage{ehrslate}{a}\ \ From EHR data to patient world models and clinical agents};

\foreach \i/\sty/\col/\ic/\head in {%
  0/stage/ehrslate/{ic sources}/{EHR sources},
  1/stagework/ehrteal/{ic layers}/{Auditable data\\infrastructure},
  2/stagefuture/ehrorange/{ic path}/{Trajectory layer},
  3/stagefuture/ehrorange/{ic agent}/{World models\\and agents},
  4/stagefuture/ehrorange/{ic evaluation}/{Evaluation}}{%
  \node[\sty, anchor=north west, minimum width=18.4\figunit,
        minimum height=15.4\figunit] at ({\i*20.4},-2.6) {};
  \pic[iconbase, draw=\col, fill=\col] at ({\i*20.4+9.2},-7.4) {\ic};
  \node[anchor=north, align=center, text width=16.6\figunit]
    at ({\i*20.4+9.2},-11.6)
    {\footnotesize\bfseries\color{\col}\nohyph\head};
}
\foreach \i in {0,1,2,3}{%
  \draw[flow] ({\i*20.4 + 18.8},-10.3) -- ({\i*20.4 + 20.0},-10.3);
}

\draw[zoom] (20.4,-18.0) -- (0.8,-21.8);
\draw[zoom] (38.8,-18.0) -- (99.2,-21.8);

\fill[detailbg] (0,-21.8) rectangle (100,-67.2);
\node[minitext, anchor=north west] at (2.4,-23.0)
  {\figstage{ehrteal}{b}\ \ Conversion, audit records, and automated validation};

\node[panel, anchor=north west, minimum width=22\figunit,
      minimum height=12.8\figunit] (src) at (2.4,-26.6) {};
\pic[iconbase, draw=ehrblue, fill=ehrblue] at (13.4,-30.8) {ic sources};
\node[anchor=north, align=center, text width=20\figunit] at (13.4,-34.4)
  {\footnotesize\bfseries\color{ehrblue}\nohyph Source layer};

\node[panel, anchor=north west, minimum width=32\figunit,
      minimum height=12.8\figunit] (can) at (30.4,-26.6) {};
\pic[iconbase, draw=ehrblue, fill=ehrblue] at (46.4,-30.8) {ic trajectory};
\node[anchor=north, align=center, text width=30\figunit] at (46.4,-34.4)
  {\footnotesize\bfseries\color{ehrblue}\nohyph Canonical events\\[0.4\figunit]
   {\scriptsize\normalfont\color{black!62}\nohyph
    occurrence\figsep availability\figsep lineage}};

\node[target, anchor=north west, minimum width=30\figunit,
      minimum height=5.8\figunit] (omop) at (67.6,-26.6) {};
\pic[iconbase, draw=ehrteal, fill=ehrteal] at (73.2,-29.5) {ic table};
\node[anchor=west] at (78.0,-29.5)
  {\footnotesize\bfseries\color{ehrteal}OMOP CDM 5.4};

\node[target, anchor=north west, minimum width=30\figunit,
      minimum height=5.8\figunit] (meds) at (67.6,-33.6) {};
\pic[iconbase, draw=ehrteal, fill=ehrteal] at (73.2,-36.5) {ic tokens};
\node[anchor=west] at (78.0,-36.5)
  {\footnotesize\bfseries\color{ehrteal}MEDS};

\draw[flow] (src.east) -- (can.west);
\draw[flow] (can.east) -- ++(2.2,0) |- (omop.west);
\draw[flow] (can.east) -- ++(2.2,0) |- (meds.west);

\fill[white, rounded corners=0.6mm] (2.4,-43.0) rectangle (97.6,-48.6);
\node[anchor=west, align=left] at (4.0,-45.8)
  {\figband{Audit records}\quad\figbody{extraction dates\figsep cohort
   membership\figsep quarantine\figsep review decisions}};

\fill[white, rounded corners=0.6mm] (2.4,-51.0) rectangle (97.6,-64.8);
\node[anchor=west, align=left] at (4.0,-53.6)
  {\figband{Validation suite}\quad\figbody{\currentCheckCount{} automated checks
   on saved outputs}};

\def\chipx{4.4}\def\chipy{-55.9}\def\chipstep{15.5}
\def\chipw{14.3\figunit}\def\chiptw{12.5\figunit}\def\chiph{7.2\figunit}
\input{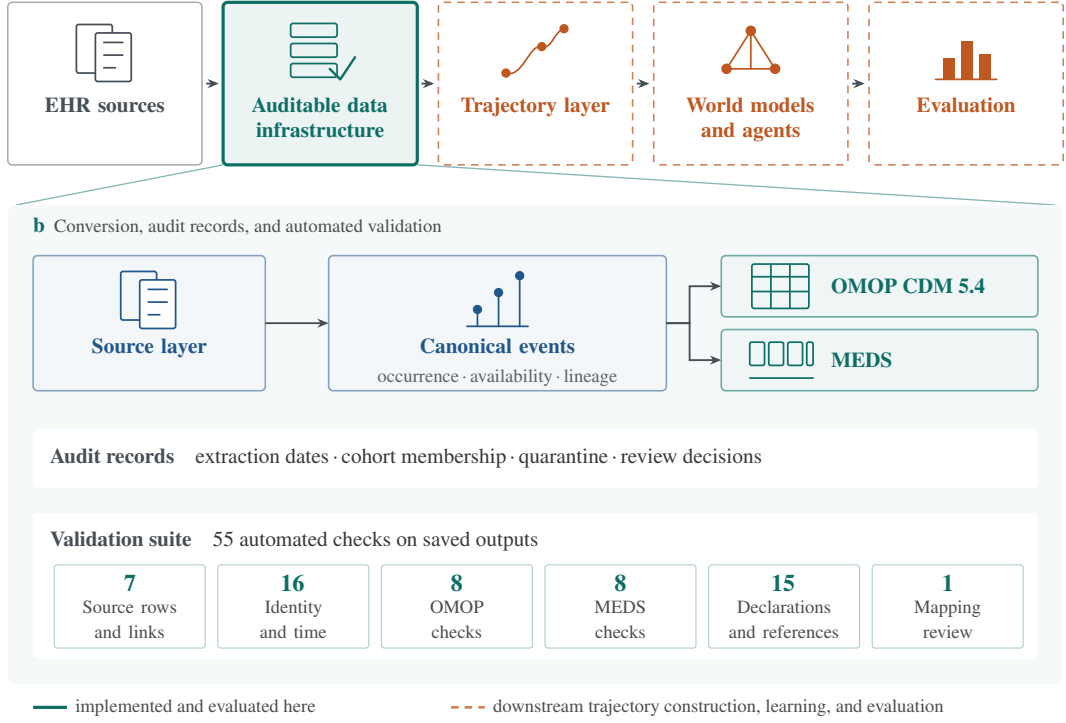}

\draw[ehrteal, line width=1.2pt] (2.4,-69.4) -- (5.6,-69.4);
\node[minitext, anchor=west] at (6.4,-69.4) {implemented and evaluated here};
\draw[ehrorange!75, line width=1.0pt, dash pattern=on 1.0mm off 0.8mm]
  (42,-69.4) -- (45.2,-69.4);
\node[minitext, anchor=west] at (46,-69.4)
  {downstream trajectory construction, learning, and evaluation};

\end{tikzpicture}
\caption{\textbf{(a)} EHR2Trace forms the data layer for patient world models and clinical agents. Dashed boxes show downstream stages. \textbf{(b)} The system converts source records into shared patient events, exports OMOP and MEDS, and stores audit records alongside the data. Automated checks validate the saved outputs. Counts show the number of checks in each group.}
\label{fig:architecture}
\end{figure}

\Needspace{4\baselineskip}
We developed EHR2Trace to preserve these distinctions during conversion and make the resulting histories auditable. The work makes three contributions:
\begin{enumerate}[itemsep=1pt]
\item \textbf{Traceable patient events.} A configurable pipeline represents source records as a shared set of patient events and exports them to OMOP and MEDS, preserving source links, timestamps, and medication action types.
\item \textbf{Automated validation.} Checks compare saved outputs with source and audit records to assess patient identity, temporal consistency, terminology mappings, and dataset splits.
\item \textbf{Empirical evaluation.} Conversions of three clinical datasets, injected faults, and repeated builds assess information preservation and reproducibility. Comparisons with published conversions characterize the information retained, and a controlled prediction experiment measures the effects of timestamp errors.
\end{enumerate}
Figure~\ref{fig:architecture} locates EHR2Trace between source data and downstream trajectory construction. The OMOP and MEDS exports integrate with existing cohort and modeling tools, with source links and temporal metadata retained for inspection.

\section{Related work}

ETHOS predicts patient trajectories from tokenized timelines, and EHRWorld studies patient states and actions for longitudinal simulation~\citep{ethos2024,ehrworld2026}. PhysicianBench evaluates clinical agents performing tasks in EHR environments~\citep{physicianbench2026}. Wang et al.\ integrate large language models with structured EHR data for predictive analytics~\citep{wang2026integrating}. Controlled evaluations of clinical vision-language models have further revealed sensitivity to workflow, prompt, and benchmark design~\citep{chase2026routing}. These applications rely on accurate records of available information and careful evaluation conditions. Healthcare reinforcement learning adds requirements for action and reward definitions, confounding control, and policy evaluation~\citep{gottesman2019}.

The OMOP Common Data Model (CDM) 5.4 organizes clinical data into tables with standard concepts. The Medical Event Data Standard (MEDS) organizes data as patient event streams~\citep{omop,omop54,meds,medsschema}. MEDS-Transforms provides preprocessing tools, ACES defines cohorts and tasks, and MEDS-DEV shares task definitions and model-training workflows~\citep{medstransforms,aces,medsdev}. EHR2Trace prepares and validates source data for use with these tools.

Kahn et al.'s framework organizes data-quality assessment around conformance, completeness, and plausibility, and the OHDSI Data Quality Dashboard (DQD) implements checks for OMOP data~\citep{kahn,dqd}. EHR2Trace extends these approaches with comparisons across source records, canonical events, and exported data, including MEDS event ordering and dataset splits. Terminology mapping relies on standard vocabularies and reviewed mappings~\citep{athena}.

\section{System design and validation}
\label{sec:system}

\subsection{Conversion with source links}

EHR2Trace separates source preservation from the representation used for export. The source layer retains raw values and row identifiers. The canonical layer represents patient events with patient and encounter identifiers, event and availability times, codes, values, reported reference ranges, and quality flags. Each event links to its source rows, whether several rows merge into one event or one row produces multiple events. Extraction dates and cohort membership are stored separately from clinical events.

Merge rules preserve distinctions that affect clinical interpretation. Records are merged only when they describe the same fact. For medication orders, dispensing, and administration, event identity includes dose, dose unit, route, status, and end time. Conflicts in other fields are resolved by a rule specified in the dataset configuration. A rule may retain the earliest value or set the field to null, flag the event, and quarantine the conflicting values. Conflicts without a declared rule are flagged and fail validation.

The OMOP exporter enforces the target model's table requirements, concept domains, and patient eligibility rules. Transfers, service changes, and intensive care unit (ICU) stays become visit details linked to a parent visit. Details without a parent visit are excluded from OMOP and logged separately. Measurement units are assigned concepts from declared unit tables and the vocabulary. Source-reported reference ranges take precedence over ranges configured for a code. Infusion rates are preserved in the drug exposure's \texttt{sig} text.

The MEDS exporter writes one code per event and retains the source code when a mapping is unavailable. Additional fields preserve availability times, event identifiers, source links, quality flags, normalized values and units, infusion rates, order actions, and discharge destinations. After identity resolution, a hash of each patient's identifier assigns that patient to a training, tuning, or held-out split. The \texttt{trace} command retrieves the source file and row associated with an exported record.

Dataset-specific conversion rules are specified in YAML files. These files define input patterns, field aliases, patient identity and merge rules, time policies, unit declarations, and terminology normalization. Adapters handle records with a single timestamp, narrative lines, measurements with multiple components, patient attributes, and visits.

Preparation scripts select columns for MIMIC-IV, CU-CTPA, and two JHU-CTPE tables, and join tables for MIMIC-IV and CU-CTPA. Row counts and input/output hashes verify this step, and each script records which delivered files it does not read. Two transformations add rows and record those additions: CU-CTPA preparation separates combined blood-pressure values into the two measurements required by OMOP, and MIMIC-IV preparation converts wide emergency department vital-sign and triage records to one row per value. Before preparation, the formats of the two JHU-CTPE tables are compared across the four cohort groups to check for export differences that could reveal cohort membership.

\subsection{Data for patient trajectories}
\label{sec:interface}

A downstream trajectory layer builds a patient history $H_t$ from these events using information available at decision time $t$. A transition takes the form
\begin{equation}
(o_t,a_t,\Delta t,o_{t+1},r_t,d_t,c_t),
\label{eq:transition}
\end{equation}
where $o_t$ is the current observation, $a_t$ the recorded action, $\Delta t$ the elapsed time, and $o_{t+1}$ the next observation. The remaining terms are a task-defined reward $r_t$, a termination flag $d_t$, and a truncation or censoring flag $c_t$.

EHR2Trace supplies the events and timestamps from which observations, actions, and elapsed times are constructed. Availability timestamps enable filtering at a decision time. Medication action types distinguish orders, dispensing, and administration, while links between these stages verify whether an order was carried out. Quality flags guide task-specific record selection, including the handling of inconsistent time intervals. The shared representation accommodates task-specific rewards and episode boundaries.

\subsection{Time, missing data, and terminology}
\label{sec:policies}

Time normalization follows a declared timezone. Death records on the same local calendar date are merged using the more precise timestamp. Records on different dates are retained and flagged, and neither is exported to the OMOP death table. Under the strict birth-year policy, patients without a derivable birth year and their dependent records are excluded from OMOP. Approximate birth years require a reference date and an approval note, and are flagged in the output. Cohort directory names remain source metadata, with labels and episodes defined downstream.

Retired source codes are mapped through relationships retained in the vocabulary. The output uses current standard concepts and flags the retired source code. Dataset configurations also define filters for nonclinical rows, such as supply containers and specimen-handling steps. Source-row accounting records these exclusions. Delivered files and columns that the conversion does not read must be declared with a reason. Records excluded from conversion are stored with reasons in a quarantine table for review.

Unit normalization preserves the source value and unit alongside any normalized values and applies only exact conversions. Corrections declared in the dataset configuration, such as interpreting Fahrenheit values under a Celsius label, require documented evidence and are flagged. Values outside the plausible range for a code and unit retain their source value, receive no normalized value, and are flagged. When a code appears in units that cannot be converted exactly, it is separated into one code per unit.

EHR2Trace stores event time and availability time in separate columns. For a timed event $e$, the inclusion rule at prediction time $\tau$ is
\begin{equation}
 t_{\mathrm{event}}(e)\leq\tau\quad\text{and}\quad
 t_{\mathrm{available}}(e)\leq\tau.
\label{eq:availability}
\end{equation}
For MIMIC-IV laboratory results, specimen collection defines event time and \texttt{storetime} defines availability time. Table~\ref{tab:trace} illustrates the filtering rule. Missing availability times are replaced by event time and marked \texttt{AVAILABILITY\_ASSUMED}, so researchers can choose whether to include these events. When duplicate records of a result have different availability times, the earliest is retained and marked \texttt{AVAILABILITY\_MERGED}.

Records without a timestamp may use a declared fallback, such as emergency department arrival time for triage measurements, and receive the flag \texttt{TIME\_FALLBACK}. Untimed attributes must appear on an approved list of stable baseline properties. Other attributes require a timestamp or are withheld. These policies define explicit, inspectable rules for selecting timed events and baseline attributes.

\begin{table}[t]
\centering\footnotesize
\begin{tabularx}{\linewidth}{@{}l r r Y l@{}}
\toprule
\texttt{event\_kind} & \multicolumn{2}{c}{hours from admission} & \texttt{source\_name}, and the detail an action carries & at $\tau$ \\
\cmidrule(lr){2-3}
 & \texttt{event} & \texttt{avail.} & & \\
\midrule
\multicolumn{5}{@{}l}{\itshape Subject 1, one encounter of 75\,h; decision time $\tau$ at +44.08\,h} \\[1pt]
\texttt{drug\_admin} & 4.37 & 4.38 & CeftriaXONE {\scriptsize\color{black!55}1 gm \textperiodcentered\ Administered} & \checkmark \\
\texttt{visit} & 7.98 & \textperiodcentered & EW EMER. {\scriptsize\color{black!55}AVAILABILITY\_ASSUMED} & \checkmark \\
\texttt{drug\_order} & 12.32 & \textperiodcentered & Vitamin D 1,000 Unit Tablet {\scriptsize\color{black!55}1000 \textperiodcentered\ PO/NG \textperiodcentered\ AVAILABILITY\_ASSUMED} & \checkmark \\
\texttt{drug\_dispense} & 12.32 & \textperiodcentered & Docusate Sodium {\scriptsize\color{black!55}PO/NG \textperiodcentered\ Discontinued via patient discharge \textperiodcentered\ AVAILABILITY\_BEFORE\_EVENT} & \checkmark \\
\texttt{drug\_order} & 12.32 & \textperiodcentered & Sodium Chloride 0.9\%  Flush 10 mL Syringe {\scriptsize\color{black!55}3 \textperiodcentered\ IV \textperiodcentered\ AVAILABILITY\_ASSUMED} & \checkmark \\
\color{ehrorange}\texttt{measurement} & 40.10 & 48.07 & Thyroid Stimulating Hormone & \textbf{withheld} \\
\color{ehrorange}\texttt{measurement} & 40.10 & 48.07 & Vitamin B12 & \textbf{withheld} \\
\multicolumn{5}{@{}l}{\quad{\scriptsize\color{black!55}261 further events in this encounter, all after $\tau$}} \\
\texttt{death} & \multicolumn{2}{r}{+5882} & death \ {\scriptsize\color{black!55}outside the encounter window} & --- \\
\midrule
\multicolumn{5}{@{}l}{\itshape Subject 2, one encounter of 2013\,h; decision time $\tau$ at +1283.29\,h} \\[1pt]
\texttt{visit} & 0.00 & \textperiodcentered & DIRECT EMER. {\scriptsize\color{black!55}AVAILABILITY\_ASSUMED} & \checkmark \\
\texttt{drug\_order} & 0.38 & \textperiodcentered & Influenza Vaccine Quadrivalent 0.5 mL Syri {\scriptsize\color{black!55}0.5 \textperiodcentered\ IM \textperiodcentered\ AVAILABILITY\_ASSUMED} & \checkmark \\
\texttt{drug\_order} & 0.38 & \textperiodcentered & PNEUMOcoccal 23-valent polysaccharide vacc {\scriptsize\color{black!55}0.5 \textperiodcentered\ IM \textperiodcentered\ AVAILABILITY\_ASSUMED} & \checkmark \\
\texttt{drug\_dispense} & 0.38 & \textperiodcentered & Sodium Chloride 0.9\%  Flush {\scriptsize\color{black!55}IV \textperiodcentered\ Discontinued \textperiodcentered\ AVAILABILITY\_BEFORE\_EVENT} & \checkmark \\
\texttt{drug\_dispense} & 0.38 & \textperiodcentered & Influenza Vaccine Quadrivalent {\scriptsize\color{black!55}IM \textperiodcentered\ Discontinued \textperiodcentered\ AVAILABILITY\_BEFORE\_EVENT} & \checkmark \\
\texttt{drug\_admin} & 1.38 & 2.52 & Sodium Chloride 0.9\%  Flush {\scriptsize\color{black!55}Not Flushed \textperiodcentered\ NOT\_ADMINISTERED} & \checkmark \\
\color{ehrorange}\texttt{measurement} & 553.87 & 2012.72 & BRONCHOALVEOLAR LAVAGE & \textbf{withheld} \\
\multicolumn{5}{@{}l}{\quad{\scriptsize\color{black!55}45648 further events in this encounter, all after $\tau$}} \\
\texttt{death} & \multicolumn{2}{r}{+772} & death \ {\scriptsize\color{black!55}outside the encounter window} & --- \\
\bottomrule
\end{tabularx}

\caption{Two patient histories from the public MIMIC-IV demonstration subset~\citep{mimicdemo}. Rows show canonical events in event-time order. A dot means event and availability times agree. Events marked ``withheld'' occurred before decision time $\tau$ but became available afterward. Orders, dispensing, and administration remain separate records, with dose, route, status, and quality flags preserved. The \texttt{NOT\_ADMINISTERED} flag records a treatment that was not given. Omitted events appear as counts.}
\label{tab:trace}
\end{table}

Terminology mapping uses reviewed mappings and vocabulary relationships. Matching after punctuation normalization requires a unique vocabulary code. When a code maps to several standard concepts, the event's domain determines which concepts are eligible for the target column. A fixed ordering resolves remaining ties, and the output records the ambiguity. Drug matching uses ingredient, strength, and dose form and records the basis for each match. Unresolved terms enter a review queue. Optional local models suggest column roles or rank supplied candidates. Human approval is required before their proposals enter the mapping registry.

\subsection{Validation of saved outputs}
\label{sec:contract}

The validation suite assesses saved outputs using \currentCheckCount{} checks. Source checks account for input rows and event-to-source links. Identity and time checks compare patient identifiers and timestamps with separately stored records. Output checks examine concept domains, keys, source links, schemas, event ordering, dataset splits, and label fields. Review checks compare model proposals with recorded decisions and accepted mappings. Each check reports its applicability, status, and diagnostic details.

Cross-layer checks assess consistency between exported data and separately stored evidence. They test whether merged source rows agree or follow a declared conflict rule, whether each source with parsed rows produces events, and whether each patient with a death event has a published death unless the dates conflict. They also compare the concepts assigned to each term in MEDS and OMOP, which use the same vocabulary.

Additional comparisons cover measurement and dose units, visit concepts, encounter links, repeated notes, excluded statuses, and the delivered files and columns. Reference tables and dataset configurations define the expected values. Configured thresholds and exceptions, such as an expected share of quarantined rows, pass validation when they satisfy their declared criteria.

Validation operates on the saved artifacts, so it detects export errors and inconsistencies in outputs generated by earlier code versions. Checks with empty or unavailable inputs are marked as skipped, and an OMOP database without patient rows is treated as absent. The report records the coverage and outcome of each validation run.

\section{Results}
\label{sec:eval}

\subsection{Conversion scale and concept coverage}
\label{sec:data}

We evaluated EHR2Trace on three datasets: a private Johns Hopkins CT pulmonary embolism extract (JHU-CTPE), MIMIC-IV v3.1 hospital and emergency data with v2.2 notes~\citep{mimic,mimic31,mimicnotes}, and a private University of Colorado CT pulmonary angiography extract (CU-CTPA). JHU-CTPE comprised four cohort partitions from two extraction batches. The MIMIC-IV configuration covered \mimicSources{} sources, and CU-CTPA included flat comma-separated files and complete clinical notes.

The conversions produced \totalCanonicalMillions{} million canonical events, with matching canonical and MEDS event counts (Table~\ref{tab:datasets}). Validation assessed the saved outputs of each dataset and reported pass, skip, and fail outcomes. All applicable checks passed for JHU-CTPE and CU-CTPA. The MIMIC-IV unit-consistency finding is detailed in Section~\ref{sec:limitations}.

\begin{table}[t]
\centering\small
\begin{tabularx}{\linewidth}{@{}Yrrr@{}}
\toprule
Measure & JHU-CTPE & MIMIC-IV & CU-CTPA \\
\midrule
Source rows & 75{,}113{,}066 & 836{,}795{,}494 & 16{,}682{,}059 \\
Identities encountered & 22{,}982 & 364{,}673 & 127{,}955 \\
Subjects with canonical events & 22{,}980 & 364{,}673 & 127{,}955 \\
Canonical events & 33{,}396{,}779 & 799{,}153{,}396 & 13{,}880{,}372 \\
Event-to-source links & 101{,}522{,}252 & 801{,}049{,}552 & 16{,}385{,}410 \\
Quarantine entries & 750{,}121 & 1{,}074{,}435 & 232{,}123 \\
OMOP persons & 22{,}980 & 364{,}627 & 127{,}955 \\
\midrule
Validation (pass / skip / fail) & 54 / 1 / 0 & 48 / 6 / 1 & 50 / 5 / 0 \\
\bottomrule
\end{tabularx}

\caption{Conversion scale and validation results. Canonical and MEDS event counts match. Patient counts differ because some identities have no events or do not meet OMOP eligibility rules. A source row can generate multiple quarantine entries.}
\label{tab:datasets}
\end{table}

Concept coverage measures the proportion of exported rows with a nonzero standard concept identifier (Table~\ref{tab:coverage}). The drug matcher mapped \drugResolved{} of \drugGold{} reviewed free-text medication terms to RxNorm using ingredient, strength, and dose form. Among these resolved terms, \drugAgreement\% matched the reviewer's concept choice.

Laboratory mappings used the community LOINC table for MIMIC-IV.\footnote{The MIT-LCP \texttt{mimic-code} file \texttt{mimic-iv/mapping/d\_labitems\_to\_loinc.csv} uses SSSOM format; \loincCommunity{} of its entries were imported after review against LOINC component and system attributes.} Spirometry, electrocardiography, and echocardiography terms in the private extracts were reviewed against the vocabulary. Configured filters removed nonclinical entries, including order-entry placeholders, supplies, and physician names in result columns.

MEDS preserves source codes and source links for events awaiting a standard concept mapping, keeping them available for source-code analyses. These records included \placeOfOccurrenceRows{} rows with ICD-10-CM place-of-occurrence codes and \atrialRateRows{} JHU-CTPE rows for 12-lead electrocardiogram atrial rate. Most other unresolved terms were medication names.

\begin{table}[t]
\centering\small
\begin{tabularx}{\linewidth}{@{}Yrrrrrr@{}}
\toprule
& \multicolumn{2}{c}{JHU-CTPE} & \multicolumn{2}{c}{MIMIC-IV} & \multicolumn{2}{c}{CU-CTPA} \\
\cmidrule(lr){2-3}\cmidrule(lr){4-5}\cmidrule(lr){6-7}
Domain & Rows & Mapped & Rows & Mapped & Rows & Mapped \\
\midrule
Condition & 3{,}637{,}437 & 99.8\% & 6{,}451{,}372 & 98.9\% & 5{,}861{,}753 & 100.0\% \\
Drug & 11{,}713{,}070 & 86.7\% & 74{,}731{,}397 & 74.5\% & 4{,}206{,}761 & 99.1\% \\
Measurement & 15{,}911{,}634 & 97.1\% & 613{,}248{,}151 & 25.7\% & 982{,}655 & 100.0\% \\
Procedure & 572{,}119 & 100.0\% & 1{,}710{,}192 & 52.6\% & 719{,}423 & 100.0\% \\
Observation & 678{,}871 & 96.6\% & 11{,}531{,}859 & 9.8\% & 35 & 100.0\% \\
\bottomrule
\end{tabularx}

\caption{OMOP concept coverage by domain, weighted by row count. ``Mapped'' indicates the presence of a standard concept identifier.}
\label{tab:coverage}
\end{table}

\subsection{Preserving time and medication actions}
\label{sec:readiness}

An audit of all canonical events confirmed that medication dispensing and administration remained distinct and that administration records retained source-provided dose and route details (Table~\ref{tab:guarantees}). The temporal layer also converted final vital status in JHU-CTPE and MIMIC-IV from an untimed attribute to a dated death event when a date was available; undated records were excluded. CU-CTPA supplied death dates directly. Table~\ref{tab:guarantees} also reports the share of assumed availability times for each dataset.

\begin{table}[t]
\centering\small
\begin{tabularx}{\linewidth}{@{}Yrrr@{}}
\toprule
Audit measure & JHU-CTPE & MIMIC-IV & CU-CTPA \\
\midrule
\multicolumn{4}{@{}l}{\itshape Vital-status timestamps} \\
Untimed \texttt{VITAL\_STATUS} events after conversion
  & \ctpeVitalTimeless{} & \mimicVitalTimeless{} & \cuVitalTimeless{} \\
\midrule
\multicolumn{4}{@{}l}{\itshape Medication action types and details} \\
Dispensing records kept separate from administration & --- & \mimicDispenses{} & --- \\
\quad cabinet records among them & --- & \mimicPyxisEvents{} & --- \\
Non-medication requests stored as service orders & --- & \mimicNonMedicationOrders{} & --- \\
Administration records with dose details & --- & \mimicEmarDose{} & --- \\
\quad with route details & --- & \mimicEmarRoute{} & --- \\
\midrule
\multicolumn{4}{@{}l}{\itshape Availability timestamps} \\
Timed events with assumed availability
  & \ctpeAssumedAvailabilityPct\% & \mimicAssumedAvailabilityPct\% & \cuAssumedAvailabilityPct\% \\
\bottomrule
\end{tabularx}

\caption{Time and medication-action audit across all canonical events. Dispensing records indicate drug supply, while administration records describe whether and how a drug was given. Assumed availability is the share of timed events that default to event time because the source lacks an availability timestamp. These events carry \texttt{AVAILABILITY\_ASSUMED}.}
\label{tab:guarantees}
\end{table}

\subsection{Comparison with published conversions}
\label{sec:comparison}

We compared EHR2Trace with published OMOP and MEDS conversions of the open MIMIC-IV demonstration subset~\citep{mimicdemoomop,mimicdemomeds}. Table~\ref{tab:comparison} reports measurements from the output files; Appendix~\ref{app:comparison} describes the procedure. EHR2Trace produced both formats from one conversion and retained the source file and row for every exported event through \cmpLineageLinks{} links. The published OMOP conversion included no source-row field, and the MEDS conversion retained source keys without their table names.

EHR2Trace stored availability time separately from event time; availability time was later than event time for \cmpAvailabilityShare{} of timed events. Neither published conversion stored a separate availability field. EHR2Trace also used an explicit field for medication orders, dispensing, and administration. The OMOP baseline used \cmpBaselineDrugTypes{} type concept for all drug rows. The MEDS baseline used one medication code, with action types recoverable from retained source keys. All three conversions read the hospital and ICU modules. The published OMOP conversion maps ICU chart items through its own item tables. EHR2Trace uses the community LOINC table for laboratory items and retains source codes for most ICU chart items. Table~\ref{tab:comparison} reports the resulting concept coverage alongside each conversion's retained information.

\begin{table}[t]
\centering\footnotesize
\begin{tabularx}{\linewidth}{@{}Y>{\raggedright\arraybackslash}p{0.24\linewidth}>{\raggedright\arraybackslash}p{0.18\linewidth}>{\raggedright\arraybackslash}p{0.19\linewidth}@{}}
\toprule
\textbf{Property} & \textbf{EHR2Trace} & \textbf{mimic-iv-demo-omop} & \textbf{mimic-iv-demo-meds} \\
\midrule
Common data models produced & OMOP and MEDS & OMOP & MEDS \\
Source modules read on the subset & hospital, ICU & hospital, ICU & hospital, ICU \\
Output rows (OMOP) / events (MEDS) & 897{,}487 / 971{,}545 & 423{,}057 & 916{,}166 \\
Rows with a standard concept & 18.9\% & 98.3\% & -- \\
Events linked to a concept & 19.8\% & -- & 13.5\% \\
Source file and row recoverable & every event (972{,}567 links) & not carried & source keys on 95.2\%, table not named \\
Availability time separate from event time & yes, later on 71.4\% of timed events & no field in the CDM & not carried \\
Medication actions distinguished & explicit field: order, dispensing, administration & one type concept for all drug rows & one code; action recoverable from source keys on 99.8\% \\
Validation report on the saved output & 54 checks, 45 passed / 8 skipped / 1 failed & 1 data-quality report & none published \\
\bottomrule
\end{tabularx}

\caption{Information retained by three conversions of the \cmpSubjects{}-patient MIMIC-IV demonstration subset. Results were measured from EHR2Trace outputs and the published baseline files. Input modules are listed for each conversion.}
\label{tab:comparison}
\end{table}

\subsection{Fault detection}
\label{sec:faults}

The validation suite detected all \faultsDetected{} faults in the injected catalogue (Figure~\ref{fig:faults}). These faults target errors that row counts, schema checks, or small-sample inspection would miss. Detection relied on comparisons with source records, audit metadata, and configuration rules. Removing the \checksAddedAfterMiss{} cross-layer checks reduced detection to \ReducedFaultsDetected{} of \faultCount{} faults across the remaining \ReducedCheckCount{} checks.

DQD detected \DqdFaultsDetected{} of the \dqdReachingWord{} injected faults that reached the OMOP database. \DqdBlindWord{} faults affected data outside OMOP and were outside DQD's scope. The cross-layer suite extends fault detection across ingest, canonical, OMOP, and MEDS artifacts.

\begin{figure}[t]
\centering
\fitwidth{\input{figures/fault_matrix}}
\caption{Detection of injected faults, grouped by the affected data layer. DQD was run on \dqdReachingWord{} faults that reach OMOP. The full suite covers all \faultCount{} faults.}
\label{fig:faults}
\end{figure}

\subsection{Effect of time errors on model evaluation}
\label{sec:downstream}

We retrospectively evaluated three temporal inclusion rules in a MIMIC-IV cohort of \leakSubjects{} patients, holding the outcome labels, patient splits, model class, and fitting procedure fixed. Rule A included timed events only when both event and availability times were at or before prediction. Rule B filtered by event time alone. Rule C assigned all diagnoses from the index admission to admission time, including diagnoses without individual timestamps. Rule C represents backdating during conversion of admission-level diagnosis tables. We fitted a separate model under each rule.

\begin{figure}[t]
\centering

\setlength{\figunit}{0.01\linewidth}%
\begin{tikzpicture}[figbase]
\begin{groupplot}[group style={group size=3 by 1, horizontal sep=10mm},
  width=0.37\linewidth, height=50mm, leakpanel,
  xtick={1,2,3,4}, xticklabels={6,12,24,48},]
\nextgroupplot[title={(a) AUROC}, ymin=0.80, ymax=1.00, ytick={0.80,0.85,0.90,0.95,1.00}, yticklabels={0.80,0.85,0.90,0.95,1.00}, legend to name=leakagelegend, legend columns=3]
\addplot[armfilter, error bars/.cd, y dir=both, y explicit] table[x=x, y=y, y error plus=ep, y error minus=em, row sep=\\] {
x y ep em \\
1 0.8294 0.0222 0.0233 \\
2 0.8389 0.0216 0.0227 \\
3 0.8536 0.0197 0.0220 \\
4 0.8629 0.0194 0.0202 \\
};
\addlegendentry{Availability filter}
\addplot[armignored, error bars/.cd, y dir=both, y explicit] table[x=x, y=y, y error plus=ep, y error minus=em, row sep=\\] {
x y ep em \\
1 0.8349 0.0209 0.0225 \\
2 0.8421 0.0214 0.0231 \\
3 0.8537 0.0197 0.0219 \\
4 0.8636 0.0191 0.0205 \\
};
\addlegendentry{Availability ignored}
\addplot[armdated, error bars/.cd, y dir=both, y explicit] table[x=x, y=y, y error plus=ep, y error minus=em, row sep=\\] {
x y ep em \\
1 0.9651 0.0088 0.0107 \\
2 0.9692 0.0082 0.0094 \\
3 0.9668 0.0081 0.0094 \\
4 0.9629 0.0095 0.0108 \\
};
\addlegendentry{Diagnosis backdating}
\nextgroupplot[title={(b) AUPRC}, ymin=0, ymax=0.80, ytick={0,0.2,0.4,0.6,0.8}, yticklabels={0.00,0.20,0.40,0.60,0.80}, xlabel={Hours after admission}]
\addplot[armfilter] coordinates {(1,0.2223) (2,0.2395) (3,0.2532) (4,0.3101)};
\addplot[armignored] coordinates {(1,0.2166) (2,0.2435) (3,0.2494) (4,0.3068)};
\addplot[armdated] coordinates {(1,0.6545) (2,0.6596) (3,0.6525) (4,0.6451)};
\nextgroupplot[title={(c) AUROC increase},
  ymode=log, ymin=0.0001, ymax=0.15, ytick={0.001,0.002,0.005,0.01,0.05,0.1},
  yticklabels={0.001,0.002,0.005,0.01,0.05,0.1},]
\addplot[armignored] coordinates {(1,0.0055) (2,0.0032) (3,0.0001) (4,0.0007)};
\addplot[armdated] coordinates {(1,0.1357) (2,0.1303) (3,0.1132) (4,0.1)};
\end{groupplot}
\end{tikzpicture}
\\[1.5mm]
\ref{leakagelegend}
\caption{Held-out performance under three time rules for the mortality-proxy task. Backdating diagnoses exposes early predictions to later information. Panel (c) shows AUROC increases relative to availability filtering on a logarithmic scale. Prediction horizons use a base-2 logarithmic axis. In panel (a), the AUROC axis begins at 0.80, and error bars show bootstrap intervals from \leakResamples{} resamples of held-out patients.}
\label{fig:leakage}
\end{figure}
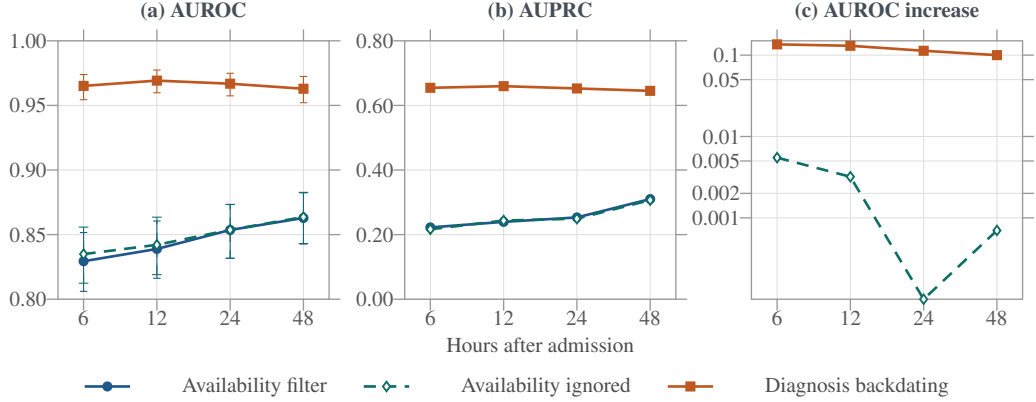

At six hours after admission, held-out AUROC was \leakCleanAuroc{} under availability filtering and \leakDatedAuroc{} under diagnosis backdating (Figure~\ref{fig:leakage}). The difference was \leakDatedGain{} (paired bootstrap interval, \leakDatedGainCi{}), estimated from \leakResamples{} resamples using the same patients for all three rules. Diagnosis backdating increased measured performance at every evaluated horizon. Filtering by event time alone had a smaller effect, indicating that the larger increase arose from assigning later admission information to earlier predictions. Appendix~\ref{app:leakage} describes the cohort and model.

Cross-rule evaluation separated the effect of the test-data rule from the effect of the training-data rule (Table~\ref{tab:leakage-transfer}). At six hours, the model trained with backdated diagnoses achieved AUROC \leakXferDatedAuroc{} on backdated test histories (C$\rightarrow$C) and \leakXferCrossAuroc{} on availability-filtered histories (C$\rightarrow$A). The model trained and evaluated with availability filtering achieved \leakXferCleanAuroc{} (A$\rightarrow$A).

For the same fitted model, changing the test rule reduced AUROC by \leakXferInflation{} (paired interval, \leakXferInflationCi{}). We refer to this contrast, C$\rightarrow$C minus C$\rightarrow$A, as evaluation inflation. Holding the test rule fixed, the model trained with backdated diagnoses underperformed the availability-filtered model by \leakXferDamage{} (paired interval, \leakXferDamageCi{}). This contrast, A$\rightarrow$A minus C$\rightarrow$A, measures the performance deficit associated with backdated training histories. The within-rule AUROC increase equals evaluation inflation minus the training deficit.

AUPRC showed the same pattern at an outcome prevalence of \leakPrevalence{}. It was \leakXferCrossAuprc{} for C$\rightarrow$A, compared with \leakCleanAuprc{} for A$\rightarrow$A and \leakDatedAuprc{} for C$\rightarrow$C. Across the \leakXferHorizons{} horizons, the AUROC deficit under a common availability-filtered test rule ranged from \leakXferDamageMin{} to \leakXferDamageMax{}. The model trained using event time alone performed within \leakXferBDamageMax{} AUROC of the availability-trained model when both were evaluated with availability filtering.

\begin{table}[t]
\centering
\caption{Retrospective evaluation under matched and mismatched training--test rules. A denotes availability filtering and C diagnosis backdating. Arrows indicate training$\rightarrow$test rules. Performance columns report held-out AUROC, with paired percentile bootstrap intervals from \leakResamples{} resamples for the two contrasts. The final column reports AUPRC for A$\rightarrow$A and C$\rightarrow$A.}
\label{tab:leakage-transfer}
\fitwidth{
\begingroup\small
\begin{tabular}{lcccccc}
\toprule
Horizon & A$\rightarrow$A & C$\rightarrow$C & C$\rightarrow$A & Evaluation inflation & Training deficit & AUPRC \\
 & filtered & backdated & cross-rule & C$\rightarrow$C $-$ C$\rightarrow$A & A$\rightarrow$A $-$ C$\rightarrow$A & A$\rightarrow$A / C$\rightarrow$A \\
\midrule
6\,h & 0.829 & 0.965 & 0.642 & 0.323 [0.295, 0.352] & 0.188 [0.161, 0.215] & 0.222 / 0.071 \\
12\,h & 0.839 & 0.969 & 0.660 & 0.309 [0.280, 0.338] & 0.179 [0.152, 0.206] & 0.239 / 0.088 \\
24\,h & 0.854 & 0.967 & 0.669 & 0.298 [0.269, 0.328] & 0.185 [0.156, 0.214] & 0.253 / 0.107 \\
48\,h & 0.863 & 0.963 & 0.694 & 0.269 [0.241, 0.297] & 0.169 [0.145, 0.194] & 0.310 / 0.142 \\
\bottomrule
\end{tabular}
\endgroup
}
\end{table}

\subsection{Reproducibility and runtime}
\label{sec:repro}

Reproducible builds rely on fixed merge ordering and output paths derived from input hashes, configuration, and code version. We tested \reproConfigs{} builds of a synthetic dataset while varying worker count, input order, and cache reuse. All builds produced \reproArtifacts{} byte-identical artifacts and passed validation. A stage-by-stage rebuild of MIMIC-IV from saved inputs reproduced the identity map, canonical events, MEDS outputs, and clinical OMOP tables.

Runtime was measured one stage at a time on an otherwise idle machine with 48 CPU cores and 251\,GB of memory. Ingestion and canonical conversion used 12 worker processes. The MIMIC-IV build required \mimicIngestMinutes{} minutes for ingestion, \mimicCanonicalMinutes{} minutes for canonical conversion, \mimicOmopMinutes{} minutes for OMOP export, and \mimicMedsMinutes{} minutes to write \mimicShards{} MEDS shards. Peak memory use during canonical conversion was \mimicCanonicalPeakGb{}\,GB.

\section{Discussion}

EHR2Trace exposes every conversion decision along the path from source records to model inputs. The shared event representation preserves clinical timing, medication actions, and source provenance across OMOP and MEDS exports. Researchers can inspect an exported event, recover its source rows, and examine the rules that produced it. This continuity enables reuse of the same clinical records across cohort construction, trajectory modeling, and agent evaluation.

The temporal experiment demonstrates the value of explicit availability rules. Under a common availability-filtered test representation, the model trained with availability filtering achieved higher AUROC and AUPRC than the model trained with backdated diagnoses. Cross-rule evaluation also quantified the inflation introduced by backdated test histories. Together, these comparisons demonstrate that temporal metadata directly improves history construction and reveals misleading performance estimates.

For patient world models of the form $p_\theta(o_{t+1},\Delta t \mid H_t,a_t)$, EHR2Trace provides explicit records for constructing the history $H_t$ and action $a_t$. Availability times enable filtering at each decision, while medication action types distinguish orders, dispensing, and administration. Source links and quality flags let researchers inspect and vary these choices. Separating conversion from task construction enables direct comparison of episode definitions, rewards, time rules, and action detail on a shared event base.

Validation and reproducibility make these representations directly auditable and reusable. Comparisons with source rows, reference tables, dataset declarations, and repeated builds verify the artifacts used in analysis. The injected-fault results demonstrate the coverage added by cross-layer checks, and the repeated builds establish reproducibility under changes in execution settings. Together with recorded conversion decisions, these capabilities enable inspection of data preparation at the level of individual events and complete datasets.

\paragraph{Broader impact.}
Source-linked histories and explicit temporal rules enable auditable training data and reproducible evaluation of clinical models. Shared exports allow researchers to reuse existing cohort and modeling tools, while recorded source links and conversion decisions provide an inspectable account of how clinical records were processed.

\Needspace{24\baselineskip}
\section{Limitations and future work}
\label{sec:limitations}

\paragraph{Evaluation scope.}
The current evaluation spans three clinical datasets, a systematic fault catalogue, a reviewed medication reference set, and a retrospective mortality-proxy task with logistic regression. Independent error collections, additional datasets, and evaluations with patient world models, clinical agents, and prospective deployment are natural next steps. Treatment-effect estimation and policy learning also require methods that address confounding and censoring~\citep{gottesman2019}.

\paragraph{Source and mapping coverage.}
Validation establishes consistency with declared rules and reference information; source completeness and clinical interpretation remain the responsibility of data owners and domain experts. Certain configurations, such as a CU-CTPA temperature column interpreted as Fahrenheit based on value ranges despite its Celsius label, reflect documented interpretations of ambiguous source metadata. The MIMIC-IV full and demonstration conversions each fail \texttt{UNIT\_HOMOGENEOUS\_}\allowbreak\texttt{PER\_CODE}. In the full dataset, twelve laboratory codes mix unit families, which manual inspection attributed to notation differences for the same quantity. Broader terminology coverage and unit harmonization are priorities for further development. Reproducibility excludes the build-date field in OMOP's \texttt{cdm\_source} metadata table.

\paragraph{Temporal coverage and use.}
Availability is assumed when the source provides no timestamp, including all timed CU-CTPA events. Individual-field availability and revisions are not tracked, and untimed attributes or interval ends may extend beyond a prediction cutoff. Field-level timing, revision tracking, and task-specific history construction will extend temporal filtering. Deployment at scale requires compliance with applicable consent, data-use, and de-identification requirements.

\section{Conclusion}

EHR2Trace delivers auditable EHR infrastructure for patient world models and clinical agents. Its shared event representation preserves source provenance, temporal metadata, and medication action types across OMOP and MEDS exports. Conversion of \totalCanonicalMillions{} million events across three clinical datasets, detection of all \faultsDetected{} injected faults, and reproducible builds demonstrate its scale and validation capabilities. The temporal experiment demonstrates that explicit availability rules improve model training and expose inflated evaluation. These capabilities make EHR preparation inspectable and reusable across downstream modeling tasks.

\ifanon
\paragraph{Ethics approval.}
The two private extracts were collected under institutional review board protocols at the two contributing institutions, which approved this study; protocol identifiers are withheld for review.
\else
\paragraph{Ethics approval.}
The JHU-CTPE extract was collected at Johns Hopkins under Institutional Review Board protocol IRB00424745 (``Radiology foundation model for chest and cardiac image interpretation''), which approved this study. The authors who collected the extract were then affiliated with Johns Hopkins. The CU-CTPA extract was collected at the University of Colorado under Institutional Review Board protocol 24-1825.
\fi

\ifanon
\paragraph{Data and code.}
EHR2Trace is released under Apache-2.0; the repository is provided as supplementary material for review and includes dataset configurations, synthetic test data, validation checks, fault and reproducibility tests used in continuous integration, and the scripts and recorded results of every experiment in this paper under \texttt{experiments/}. JHU-CTPE and CU-CTPA are private. MIMIC-IV access follows PhysioNet credentialing and data-use conditions~\citep{mimic31,mimicnotes,physionet2026}. Patient-level outputs and dataset-specific run records are not redistributed. Clinical vocabularies are obtained through Athena under the terms of their respective owners~\citep{athena}.
\else
\paragraph{Data and code.}
EHR2Trace is available under Apache-2.0 at \url{https://github.com/Yangxinyee/ehr2trace}, including dataset configurations, synthetic test data, validation checks, fault and reproducibility tests used in continuous integration, and the scripts and recorded results of every experiment in this paper under \texttt{experiments/}. JHU-CTPE and CU-CTPA are private. MIMIC-IV access follows PhysioNet credentialing and data-use conditions~\citep{mimic31,mimicnotes,physionet2026}. Patient-level outputs and dataset-specific run records are not redistributed. Clinical vocabularies are obtained through Athena under the terms of their respective owners~\citep{athena}.
\fi

\paragraph{Writing assistance.}
OpenAI Codex and Anthropic Claude assisted with language editing, consistency checks, and figure-generation code.

\begingroup
\small
\setlength{\bibsep}{4pt}
\bibliographystyle{plainnat}
\bibliography{refs}
\endgroup

\clearpage
\appendix
\section{Converter comparison details}
\label{app:comparison}

We used the MIMIC-IV demonstration release of \cmpSubjects{} patients, available under the Open Data Commons Open Database License. The OMOP baseline was \texttt{mimic-iv-demo-omop} v0.9~\citep{mimicdemoomop}, with OMOP CDM 5.3.1, a September 2020 vocabulary release, and \cmpBaselineOmopRows{} clinical rows. It included an OHDSI Data Quality Dashboard report. The MEDS baseline was \texttt{mimic-iv-demo-meds} v0.0.1~\citep{mimicdemomeds}, with \cmpBaselineMedsEvents{} events in MEDS 0.3.3 format, produced by MEDS-Transforms. We measured both baselines from their published files. EHR2Trace used the full-dataset configuration on the same subset and produced \cmpOurEvents{} events. This conversion also runs in continuous integration on every code change and is reproducible from the repository.

Table~\ref{tab:comparison} measures source recovery as the presence of both a source table name and a row identifier. Availability requires a separate timestamp; the reported share is the proportion of timed events for which availability time is later than event time. Medication action types are the values used to distinguish ordering, dispensing, and administration. The OMOP baseline records \cmpBaselineDrugTypes{} drug type concept. The MEDS baseline uses one medication code and retains order and administration keys on \cmpBaselineMedRecoverable{} of medication events.

Concept coverage uses each model's own denominator: clinical rows for OMOP and events for MEDS. We report source modules alongside output counts because the modules determine which tables and events are included. Counts and coverage should be compared within matching modules. Concept coverage also depends on the vocabulary release.

The script \texttt{experiments/run\_converter\_}\allowbreak\texttt{comparison.py} in the software repository writes the measurements to \texttt{experiments/results/}\allowbreak\texttt{converter\_comparison.json}; the table is generated from that record.

\section{Temporal experiment details}
\label{app:leakage}

The cohort included \leakSubjects{} MIMIC-IV patients, each represented by the first qualifying admission lasting at least 72 hours. A patient received a positive mortality-proxy label if the earliest recorded death date was no later than one day after discharge. This yielded \leakDeaths{} positive cases (\leakPrevalence{}). Features were log-transformed code counts, including prior history and untimed attributes. We used logistic regression with an $\ell_2$ penalty, \texttt{liblinear}, $C=1$, a maximum of 2,000 iterations, and seed 20260826. The training and held-out sets contained 104,858 and 13,052 patients, respectively. The tuning set was unused.

Each arm built its code index and fitted its model using only its training split. Codes found only in held-out patients were excluded from the feature set. We calculated percentile bootstrap intervals from \leakResamples{} resamples of held-out patients. Each resample used the same patients for every arm, yielding paired estimates of the AUROC increase relative to availability filtering.

For cross-rule evaluation, each fitted model retained the code vocabulary of its training rule. Test histories were represented in that vocabulary, with zeros for codes absent under the test rule. All nine training--test combinations at a given horizon were evaluated using the same bootstrap resamples, yielding paired estimates for the reported contrasts. Results for matched training and test rules agreed with the single-rule experiments to within 0.001. A mismatch in the other direction also reduced performance: the model trained with availability filtering achieved AUROC \leakXferReverseAuroc{} on backdated test histories (A$\rightarrow$C) at six hours. This combination corresponds to no realistic deployment scenario and is omitted from Table~\ref{tab:leakage-transfer}. The training deficit compares the two models under the availability-filtered rule, the condition under which either would be deployed.

\ifanon
\newpage
\section*{NeurIPS Paper Checklist}

\begin{enumerate}

\item {\bf Claims}
    \item[] Question: Do the main claims made in the abstract and introduction accurately reflect the paper's contributions and scope?
    \item[] Answer: \answerYes{} 
    \item[] Justification: The abstract and introduction state the three contributions (traceable events with OMOP and MEDS exports, automated validation, and the evaluation on three datasets) and report the measured results at the scope of those datasets; the limitations section bounds the claims.
    \item[] Guidelines:
    \begin{itemize}
        \item The answer \answerNA{} means that the abstract and introduction do not include the claims made in the paper.
        \item The abstract and/or introduction should clearly state the claims made, including the contributions made in the paper and important assumptions and limitations. A \answerNo{} or \answerNA{} answer to this question will not be perceived well by the reviewers. 
        \item The claims made should match theoretical and experimental results, and reflect how much the results can be expected to generalize to other settings. 
        \item It is fine to include aspirational goals as motivation as long as it is clear that these goals are not attained by the paper. 
    \end{itemize}

\item {\bf Limitations}
    \item[] Question: Does the paper discuss the limitations of the work performed by the authors?
    \item[] Answer: \answerYes{} 
    \item[] Justification: Section~\ref{sec:limitations} consolidates evaluation scope, source and mapping coverage, temporal coverage, and future work, including source-interpretation uncertainty and the recorded unit-consistency finding.
    \item[] Guidelines:
    \begin{itemize}
        \item The answer \answerNA{} means that the paper has no limitation while the answer \answerNo{} means that the paper has limitations, but those are not discussed in the paper. 
        \item The authors are encouraged to create a separate ``Limitations'' section in their paper.
        \item The paper should point out any strong assumptions and how robust the results are to violations of these assumptions (e.g., independence assumptions, noiseless settings, model well-specification, asymptotic approximations only holding locally). The authors should reflect on how these assumptions might be violated in practice and what the implications would be.
        \item The authors should reflect on the scope of the claims made, e.g., if the approach was only tested on a few datasets or with a few runs. In general, empirical results often depend on implicit assumptions, which should be articulated.
        \item The authors should reflect on the factors that influence the performance of the approach. For example, a facial recognition algorithm may perform poorly when image resolution is low or images are taken in low lighting. Or a speech-to-text system might not be used reliably to provide closed captions for online lectures because it fails to handle technical jargon.
        \item The authors should discuss the computational efficiency of the proposed algorithms and how they scale with dataset size.
        \item If applicable, the authors should discuss possible limitations of their approach to address problems of privacy and fairness.
        \item While the authors might fear that complete honesty about limitations might be used by reviewers as grounds for rejection, a worse outcome might be that reviewers discover limitations that aren't acknowledged in the paper. The authors should use their best judgment and recognize that individual actions in favor of transparency play an important role in developing norms that preserve the integrity of the community. Reviewers will be specifically instructed to not penalize honesty concerning limitations.
    \end{itemize}

\item {\bf Theory assumptions and proofs}
    \item[] Question: For each theoretical result, does the paper provide the full set of assumptions and a complete (and correct) proof?
    \item[] Answer: \answerNA{} 
    \item[] Justification: The paper contains no theoretical results.
    \item[] Guidelines:
    \begin{itemize}
        \item The answer \answerNA{} means that the paper does not include theoretical results. 
        \item All the theorems, formulas, and proofs in the paper should be numbered and cross-referenced.
        \item All assumptions should be clearly stated or referenced in the statement of any theorems.
        \item The proofs can either appear in the main paper or the supplemental material, but if they appear in the supplemental material, the authors are encouraged to provide a short proof sketch to provide intuition. 
        \item Inversely, any informal proof provided in the core of the paper should be complemented by formal proofs provided in appendix or supplemental material.
        \item Theorems and Lemmas that the proof relies upon should be properly referenced. 
    \end{itemize}

    \item {\bf Experimental result reproducibility}
    \item[] Question: Does the paper fully disclose all the information needed to reproduce the main experimental results of the paper to the extent that it affects the main claims and/or conclusions of the paper (regardless of whether the code and data are provided or not)?
    \item[] Answer: \answerYes{} 
    \item[] Justification: The pipeline, dataset configurations, validation checks, fault catalogue, and reproducibility tests are in the released repository; Appendix~\ref{app:comparison} and Appendix~\ref{app:leakage} give the comparison procedure and the temporal experiment's cohort, features, model, hyperparameters, splits, and seed. MIMIC-IV is available to credentialed PhysioNet users; the two private extracts cannot be shared.
    \item[] Guidelines:
    \begin{itemize}
        \item The answer \answerNA{} means that the paper does not include experiments.
        \item If the paper includes experiments, a \answerNo{} answer to this question will not be perceived well by the reviewers: Making the paper reproducible is important, regardless of whether the code and data are provided or not.
        \item If the contribution is a dataset and\slash or model, the authors should describe the steps taken to make their results reproducible or verifiable. 
        \item Depending on the contribution, reproducibility can be accomplished in various ways. For example, if the contribution is a novel architecture, describing the architecture fully might suffice, or if the contribution is a specific model and empirical evaluation, it may be necessary to either make it possible for others to replicate the model with the same dataset, or provide access to the model. In general. releasing code and data is often one good way to accomplish this, but reproducibility can also be provided via detailed instructions for how to replicate the results, access to a hosted model (e.g., in the case of a large language model), releasing of a model checkpoint, or other means that are appropriate to the research performed.
        \item While NeurIPS does not require releasing code, the conference does require all submissions to provide some reasonable avenue for reproducibility, which may depend on the nature of the contribution. For example
        \begin{enumerate}
            \item If the contribution is primarily a new algorithm, the paper should make it clear how to reproduce that algorithm.
            \item If the contribution is primarily a new model architecture, the paper should describe the architecture clearly and fully.
            \item If the contribution is a new model (e.g., a large language model), then there should either be a way to access this model for reproducing the results or a way to reproduce the model (e.g., with an open-source dataset or instructions for how to construct the dataset).
            \item We recognize that reproducibility may be tricky in some cases, in which case authors are welcome to describe the particular way they provide for reproducibility. In the case of closed-source models, it may be that access to the model is limited in some way (e.g., to registered users), but it should be possible for other researchers to have some path to reproducing or verifying the results.
        \end{enumerate}
    \end{itemize}

\item {\bf Open access to data and code}
    \item[] Question: Does the paper provide open access to the data and code, with sufficient instructions to faithfully reproduce the main experimental results, as described in supplemental material?
    \item[] Answer: \answerYes{} 
    \item[] Justification: Code, configurations, synthetic test data, and the experiment scripts are released under Apache-2.0 with build instructions. MIMIC-IV requires PhysioNet credentialing; JHU-CTPE and CU-CTPA are private, and patient-level outputs are not redistributed.
    \item[] Guidelines:
    \begin{itemize}
        \item The answer \answerNA{} means that paper does not include experiments requiring code.
        \item Please see the NeurIPS code and data submission guidelines (\url{https://neurips.cc/public/guides/CodeSubmissionPolicy}) for more details.
        \item While we encourage the release of code and data, we understand that this might not be possible, so \answerNo{} is an acceptable answer. Papers cannot be rejected simply for not including code, unless this is central to the contribution (e.g., for a new open-source benchmark).
        \item The instructions should contain the exact command and environment needed to run to reproduce the results. See the NeurIPS code and data submission guidelines (\url{https://neurips.cc/public/guides/CodeSubmissionPolicy}) for more details.
        \item The authors should provide instructions on data access and preparation, including how to access the raw data, preprocessed data, intermediate data, and generated data, etc.
        \item The authors should provide scripts to reproduce all experimental results for the new proposed method and baselines. If only a subset of experiments are reproducible, they should state which ones are omitted from the script and why.
        \item At submission time, to preserve anonymity, the authors should release anonymized versions (if applicable).
        \item Providing as much information as possible in supplemental material (appended to the paper) is recommended, but including URLs to data and code is permitted.
    \end{itemize}

\item {\bf Experimental setting/details}
    \item[] Question: Does the paper specify all the training and test details (e.g., data splits, hyperparameters, how they were chosen, type of optimizer) necessary to understand the results?
    \item[] Answer: \answerYes{} 
    \item[] Justification: Appendix~\ref{app:leakage} specifies the cohort definition, label, features, model, regularization, solver, iteration cap, seed, and the sizes of the training and held-out sets; data splits are assigned by a hash of the patient identifier as described in Section~\ref{sec:system}.
    \item[] Guidelines:
    \begin{itemize}
        \item The answer \answerNA{} means that the paper does not include experiments.
        \item The experimental setting should be presented in the core of the paper to a level of detail that is necessary to appreciate the results and make sense of them.
        \item The full details can be provided either with the code, in appendix, or as supplemental material.
    \end{itemize}

\item {\bf Experiment statistical significance}
    \item[] Question: Does the paper report error bars suitably and correctly defined or other appropriate information about the statistical significance of the experiments?
    \item[] Answer: \answerYes{} 
    \item[] Justification: Section~\ref{sec:downstream} and Appendix~\ref{app:leakage} report paired percentile bootstrap intervals over \leakResamples{} resamples of the held-out patients; the factor of variability is the held-out patient sample, and the intervals in Table~\ref{tab:leakage-transfer} and Figure~\ref{fig:leakage} are computed from the same resamples.
    \item[] Guidelines:
    \begin{itemize}
        \item The answer \answerNA{} means that the paper does not include experiments.
        \item The authors should answer \answerYes{} if the results are accompanied by error bars, confidence intervals, or statistical significance tests, at least for the experiments that support the main claims of the paper.
        \item The factors of variability that the error bars are capturing should be clearly stated (for example, train/test split, initialization, random drawing of some parameter, or overall run with given experimental conditions).
        \item The method for calculating the error bars should be explained (closed form formula, call to a library function, bootstrap, etc.)
        \item The assumptions made should be given (e.g., Normally distributed errors).
        \item It should be clear whether the error bar is the standard deviation or the standard error of the mean.
        \item It is OK to report 1-sigma error bars, but one should state it. The authors should preferably report a 2-sigma error bar than state that they have a 96\% CI, if the hypothesis of Normality of errors is not verified.
        \item For asymmetric distributions, the authors should be careful not to show in tables or figures symmetric error bars that would yield results that are out of range (e.g., negative error rates).
        \item If error bars are reported in tables or plots, the authors should explain in the text how they were calculated and reference the corresponding figures or tables in the text.
    \end{itemize}

\item {\bf Experiments compute resources}
    \item[] Question: For each experiment, does the paper provide sufficient information on the computer resources (type of compute workers, memory, time of execution) needed to reproduce the experiments?
    \item[] Answer: \answerYes{} 
    \item[] Justification: Section~\ref{sec:repro} reports the machine (48 CPU cores, 251\,GB of memory, no GPU) and the wall time and peak memory of each MIMIC-IV stage; the temporal experiment runs on the same machine in minutes. Development rebuilds used more compute than the reported runs.
    \item[] Guidelines:
    \begin{itemize}
        \item The answer \answerNA{} means that the paper does not include experiments.
        \item The paper should indicate the type of compute workers CPU or GPU, internal cluster, or cloud provider, including relevant memory and storage.
        \item The paper should provide the amount of compute required for each of the individual experimental runs as well as estimate the total compute. 
        \item The paper should disclose whether the full research project required more compute than the experiments reported in the paper (e.g., preliminary or failed experiments that didn't make it into the paper). 
    \end{itemize}
    
\item {\bf Code of ethics}
    \item[] Question: Does the research conducted in the paper conform, in every respect, with the NeurIPS Code of Ethics \url{https://neurips.cc/public/EthicsGuidelines}?
    \item[] Answer: \answerYes{} 
    \item[] Justification: The work uses de-identified records under institutional review board approval and PhysioNet data-use conditions, redistributes no patient-level data, and the conversion records every decision it makes for audit.
    \item[] Guidelines:
    \begin{itemize}
        \item The answer \answerNA{} means that the authors have not reviewed the NeurIPS Code of Ethics.
        \item If the authors answer \answerNo, they should explain the special circumstances that require a deviation from the Code of Ethics.
        \item The authors should make sure to preserve anonymity (e.g., if there is a special consideration due to laws or regulations in their jurisdiction).
    \end{itemize}

\item {\bf Broader impacts}
    \item[] Question: Does the paper discuss both potential positive societal impacts and negative societal impacts of the work performed?
    \item[] Answer: \answerYes{} 
    \item[] Justification: The Broader impact paragraph describes auditable training data, reproducible evaluation, and reuse of clinical records. Section~\ref{sec:limitations} discusses source completeness and accuracy, consent, data-use and de-identification requirements for use at scale. Source links and recorded decisions enable review of data processing.
    \item[] Guidelines:
    \begin{itemize}
        \item The answer \answerNA{} means that there is no societal impact of the work performed.
        \item If the authors answer \answerNA{} or \answerNo, they should explain why their work has no societal impact or why the paper does not address societal impact.
        \item Examples of negative societal impacts include potential malicious or unintended uses (e.g., disinformation, generating fake profiles, surveillance), fairness considerations (e.g., deployment of technologies that could make decisions that unfairly impact specific groups), privacy considerations, and security considerations.
        \item The conference expects that many papers will be foundational research and not tied to particular applications, let alone deployments. However, if there is a direct path to any negative applications, the authors should point it out. For example, it is legitimate to point out that an improvement in the quality of generative models could be used to generate Deepfakes for disinformation. On the other hand, it is not needed to point out that a generic algorithm for optimizing neural networks could enable people to train models that generate Deepfakes faster.
        \item The authors should consider possible harms that could arise when the technology is being used as intended and functioning correctly, harms that could arise when the technology is being used as intended but gives incorrect results, and harms following from (intentional or unintentional) misuse of the technology.
        \item If there are negative societal impacts, the authors could also discuss possible mitigation strategies (e.g., gated release of models, providing defenses in addition to attacks, mechanisms for monitoring misuse, mechanisms to monitor how a system learns from feedback over time, improving the efficiency and accessibility of ML).
    \end{itemize}
    
\item {\bf Safeguards}
    \item[] Question: Does the paper describe safeguards that have been put in place for responsible release of data or models that have a high risk for misuse (e.g., pre-trained language models, image generators, or scraped datasets)?
    \item[] Answer: \answerNA{} 
    \item[] Justification: The paper releases code and synthetic test data only; no model or dataset with a high risk of misuse is released.
    \item[] Guidelines:
    \begin{itemize}
        \item The answer \answerNA{} means that the paper poses no such risks.
        \item Released models that have a high risk for misuse or dual-use should be released with necessary safeguards to allow for controlled use of the model, for example by requiring that users adhere to usage guidelines or restrictions to access the model or implementing safety filters. 
        \item Datasets that have been scraped from the Internet could pose safety risks. The authors should describe how they avoided releasing unsafe images.
        \item We recognize that providing effective safeguards is challenging, and many papers do not require this, but we encourage authors to take this into account and make a best faith effort.
    \end{itemize}

\item {\bf Licenses for existing assets}
    \item[] Question: Are the creators or original owners of assets (e.g., code, data, models), used in the paper, properly credited and are the license and terms of use explicitly mentioned and properly respected?
    \item[] Answer: \answerYes{} 
    \item[] Justification: MIMIC-IV and its demonstration subset are cited with versions and used under their PhysioNet and Open Database licenses, the OMOP and MEDS baselines are cited with versions, and the vocabularies are obtained through Athena under their owners' terms.
    \item[] Guidelines:
    \begin{itemize}
        \item The answer \answerNA{} means that the paper does not use existing assets.
        \item The authors should cite the original paper that produced the code package or dataset.
        \item The authors should state which version of the asset is used and, if possible, include a URL.
        \item The name of the license (e.g., CC-BY 4.0) should be included for each asset.
        \item For scraped data from a particular source (e.g., website), the copyright and terms of service of that source should be provided.
        \item If assets are released, the license, copyright information, and terms of use in the package should be provided. For popular datasets, \url{paperswithcode.com/datasets} has curated licenses for some datasets. Their licensing guide can help determine the license of a dataset.
        \item For existing datasets that are re-packaged, both the original license and the license of the derived asset (if it has changed) should be provided.
        \item If this information is not available online, the authors are encouraged to reach out to the asset's creators.
    \end{itemize}

\item {\bf New assets}
    \item[] Question: Are new assets introduced in the paper well documented and is the documentation provided alongside the assets?
    \item[] Answer: \answerYes{} 
    \item[] Justification: The EHR2Trace repository is documented with dataset configurations, tests, and build instructions, and is released under Apache-2.0.
    \item[] Guidelines:
    \begin{itemize}
        \item The answer \answerNA{} means that the paper does not release new assets.
        \item Researchers should communicate the details of the dataset\slash code\slash model as part of their submissions via structured templates. This includes details about training, license, limitations, etc. 
        \item The paper should discuss whether and how consent was obtained from people whose asset is used.
        \item At submission time, remember to anonymize your assets (if applicable). You can either create an anonymized URL or include an anonymized zip file.
    \end{itemize}

\item {\bf Crowdsourcing and research with human subjects}
    \item[] Question: For crowdsourcing experiments and research with human subjects, does the paper include the full text of instructions given to participants and screenshots, if applicable, as well as details about compensation (if any)? 
    \item[] Answer: \answerNA{} 
    \item[] Justification: The paper involves no crowdsourcing and no contact with human participants; it converts retrospective de-identified records.
    \item[] Guidelines:
    \begin{itemize}
        \item The answer \answerNA{} means that the paper does not involve crowdsourcing nor research with human subjects.
        \item Including this information in the supplemental material is fine, but if the main contribution of the paper involves human subjects, then as much detail as possible should be included in the main paper. 
        \item According to the NeurIPS Code of Ethics, workers involved in data collection, curation, or other labor should be paid at least the minimum wage in the country of the data collector. 
    \end{itemize}

\item {\bf Institutional review board (IRB) approvals or equivalent for research with human subjects}
    \item[] Question: Does the paper describe potential risks incurred by study participants, whether such risks were disclosed to the subjects, and whether Institutional Review Board (IRB) approvals (or an equivalent approval/review based on the requirements of your country or institution) were obtained?
    \item[] Answer: \answerYes{} 
    \item[] Justification: The ethics approval paragraph names the institutional review board protocols under which the two private extracts were collected and used; MIMIC-IV is used under PhysioNet's credentialing. Participants incur no new risk from retrospective record conversion.
    \item[] Guidelines:
    \begin{itemize}
        \item The answer \answerNA{} means that the paper does not involve crowdsourcing nor research with human subjects.
        \item Depending on the country in which research is conducted, IRB approval (or equivalent) may be required for any human subjects research. If you obtained IRB approval, you should clearly state this in the paper. 
        \item We recognize that the procedures for this may vary significantly between institutions and locations, and we expect authors to adhere to the NeurIPS Code of Ethics and the guidelines for their institution. 
        \item For initial submissions, do not include any information that would break anonymity (if applicable), such as the institution conducting the review.
    \end{itemize}

\item {\bf Declaration of LLM usage}
    \item[] Question: Does the paper describe the usage of LLMs if it is an important, original, or non-standard component of the core methods in this research? Note that if the LLM is used only for writing, editing, or formatting purposes and does \emph{not} impact the core methodology, scientific rigor, or originality of the research, declaration is not required.
    \item[] Answer: \answerYes{} 
    \item[] Justification: Section~\ref{sec:policies} describes the optional local models that suggest column roles or rank candidate mappings, whose proposals require human approval; the writing-assistance paragraph declares the LLMs used for language editing and figure code.
    \item[] Guidelines:
    \begin{itemize}
        \item The answer \answerNA{} means that the core method development in this research does not involve LLMs as any important, original, or non-standard components.
        \item Please refer to our LLM policy in the NeurIPS handbook for what should or should not be described.
    \end{itemize}

\end{enumerate}
\fi

\end{document}